\documentclass[11pt]{article}
\usepackage{acl2013}
\usepackage{times}
\usepackage{url}
\usepackage{latexsym}
\usepackage{graphicx}
\usepackage{amsmath}
\renewcommand{\vec}[1]{\mathbf{#1}} 

\title{Recurrent Convolutional Neural Networks for Discourse Compositionality}

\author{Nal Kalchbrenner \\
  Department of Computer Science \\
  Oxford University \\
  {\tt nkalch@cs.ox.ac.uk} \\\And
  Phil Blunsom \\
  Department of Computer Science  \\
  Oxford University \\
  {\tt pblunsom@cs.ox.ac.uk} \\}
\date{}

\begin{document}
\maketitle
\begin{abstract}
The compositionality of meaning extends beyond the single sentence. Just as words combine to form the meaning of sentences, so do sentences combine to form the meaning of paragraphs, dialogues and general discourse. We introduce both a sentence model and a discourse model corresponding to the two levels of compositionality. The sentence model adopts convolution as the central operation for composing semantic vectors and is based on a novel hierarchical convolutional neural network. The discourse model extends the sentence model and is based on a recurrent neural network that is conditioned in a novel way both on the current sentence and on the current speaker. The discourse model is able to capture both the sequentiality of sentences and the interaction between different speakers. Without feature engineering or pretraining and with simple greedy decoding, the discourse model coupled to the sentence model obtains state of the art performance on a dialogue act classification experiment.

\end{abstract}

\section{Introduction}

There are at least two levels at which the meaning of smaller linguistic units is composed to form the meaning of larger linguistic units. The first level is that of sentential compositionality, where the meaning of words composes to form the meaning of the sentence or utterance that contains them \cite{Frege}. The second level extends beyond the first and involves general discourse compositionality, where the meaning of multiple sentences or utterances composes to form the meaning of the paragraph, document or dialogue that comprises them \cite{sep-pragmatics,Potts2011PRAGMATICS}. The problem of discourse compositionality is the problem of modelling how the meaning of general discourse composes from the meaning of the sentences involved and, since the latter in turn stems from the meaning of the words, how the meaning of discourse composes from the words themselves.

Tackling the problem of discourse compositionality promises to be central to a number of different applications. These include sentiment or topic classification of single sentences within the context of a longer discourse, the recognition of dialogue acts within a conversation, the classification of a discourse as a whole and the attainment of general unsupervised or semi-supervised representations of a discourse for potential use in dialogue  tracking and question answering systems and machine translation, among others.

To this end much work has been done on modelling the meaning of single words by way of semantic vectors \cite{DBLP:journals/jair/TurneyP10,collobert:2008} and the latter have found applicability  in areas such as information retrieval \cite{DBLP:conf/www/JonesRMG06}. With regard to modelling the meaning of sentences and sentential compositionality, recent proposals have included simple additive and multiplicative models that do not take into account sentential features such as word order or syntactic structure \cite{DBLP:journals/cogsci/MitchellL10}, matrix-vector based models that do take into account such features but are limited to phrases of a specific syntactic type \cite{DBLP:conf/emnlp/BaroniZ10} and  structured models that fully capture such  features \cite{DBLP:journals/corr/abs-1101-0309} and are embedded within a deep neural architecture \cite{SocherEtAl2012:MVRNN,hermann-blunsom:2013:ACL2013}. It is notable that the additive and multiplicative models as well as simple, non-compositional bag of $n$-grams and word vector averaging models have  equalled or outperformed the structured models at certain phrase similarity \cite{DBLP:conf/emnlp/BlacoeL12} and sentiment classification tasks \cite{DBLP:journals/corr/abs-1301-2811,DBLP:conf/acl/WangM12}.

With regard to discourse compositionality, most of the proposals aimed at capturing semantic aspects of paragraphs or longer texts have focused on bag of $n$-grams or sentence vector averaging approaches \cite{DBLP:conf/acl/WangM12,SocherEtAl2012:MVRNN}.  In addition, the recognition of dialogue acts within dialogues has largely been treated in non-compositional ways by way of language models coupled to hidden Markov sequence models \cite{DBLP:journals/coling/StolckeRCSBJTMM00}.  Principled approaches to discourse compositionality have largely been unexplored.

We introduce a novel model for sentential compositionality. The composition operation is based on a hierarchy of one dimensional convolutions. The convolutions are applied feature-wise, that is they are applied across each feature of the word vectors in the sentence. The weights adopted in each convolution are different for each feature, but do not depend on the different words being composed. The hierarchy of convolution operations involves a sequence of convolution kernels of increasing sizes (Fig.\,1). This allows for the composition operation to be applied to sentences of any length, while keeping the model at a depth of roughly $\sqrt{2l}$ where $l$ is the length of the sentence. The hierarchy of feature-wise convolution operations followed by sigmoid non-linear activation functions results in a hierarchical convolutional neural network (HCNN) based on a convolutional architecture \cite{lecun-01a}.
The HCNN shares with the structured models the aspect that it is sensitive to word order and adopts a hierarchical architecture, although it is not based on explicit syntactic structure. 
\begin{figure}
\includegraphics[width=0.45\textwidth]{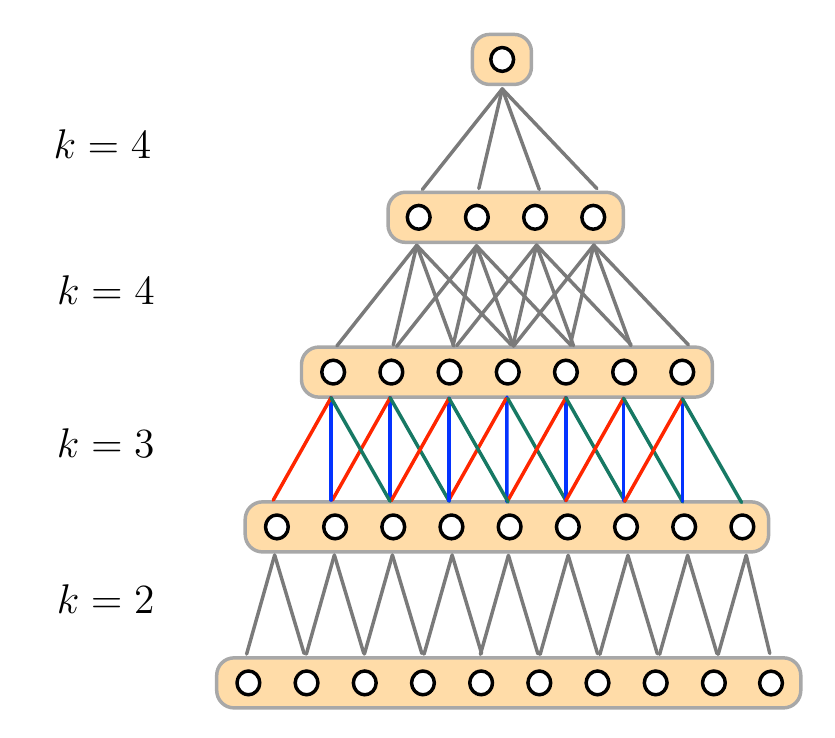}

\caption{A hierarchical convolutional neural network for sentential compositionality. The bottom layer represents a single feature across all the word vectors in the sentence. The top layer is the value for that feature in the resulting sentence vector. Lines represent single weights and color coded lines indicate sharing of weights. The parameter $k$ indicates the size of the convolution kernel at the corresponding layer.}
\end{figure}
 
We also introduce a novel model for discourse compositionality. The discourse model is based on a recurrent neural network (RNN) architecture that is a powerful model for sequences \cite{DBLP:conf/icml/SutskeverMH11,DBLP:conf/interspeech/MikolovKBCK10}. The model aims at capturing two central aspects of discourse and its meaning: the sequentiality of the sentences or utterances in the discourse and, where applicable, the interactions between the different speakers. The underlying RNN has its recurrent and output weights conditioned on the respective speaker, while simultaneously taking as input at every turn the sentence vector for the current sentence generated through the sentence model  (Fig.\,2).

We experiment with the discourse model coupled to the sentence model on the task of recognizing dialogue acts of utterances within a conversation. The dataset is given by 1134 transcribed and annotated telephone conversations amounting to about 200K utterances from the Switchboard Dialogue Act Corpus \cite{0a0045064fb44dcc957522a58fd75771}.\footnote{The dataset is available at \texttt{compprag. christopherpotts.net/swda.html}}
The model is trained in a supervised setting without previous pretraining; word vectors are also randomly initialised. The model learns a probability distribution over the dialogue acts at step $i$ given the sequence of utterances up to step $i$, the sequence of acts up to the previous step $i-1$ and the binary sequence of agents up to the current step $i$. Predicting the sequence of dialogue acts is performed in a greedy fashion.\footnote{Code and trained model available at \texttt{nal.co}}

We proceed as follows. In Sect. 2 we give the motivation and the definition for the HCNN sentence model. In Sect. 3 we do the same for the RCNN discourse model. In Sect. 4 we describe the dialogue act classification experiment and the training procedure. We also inspect the discourse vector representations produced by the model. We conclude in Sect. 5.

\section{Sentence Model}

The general aim of the sentence model is to compute a vector for a sentence $s$ given the sequence of words in $s$ and a vector for each of the words. The computation captures certain general considerations regarding sentential compositionality. We first relate such considerations and we then proceed to give a definition of the model. 
\begin{figure}
\includegraphics[width=0.48\textwidth]{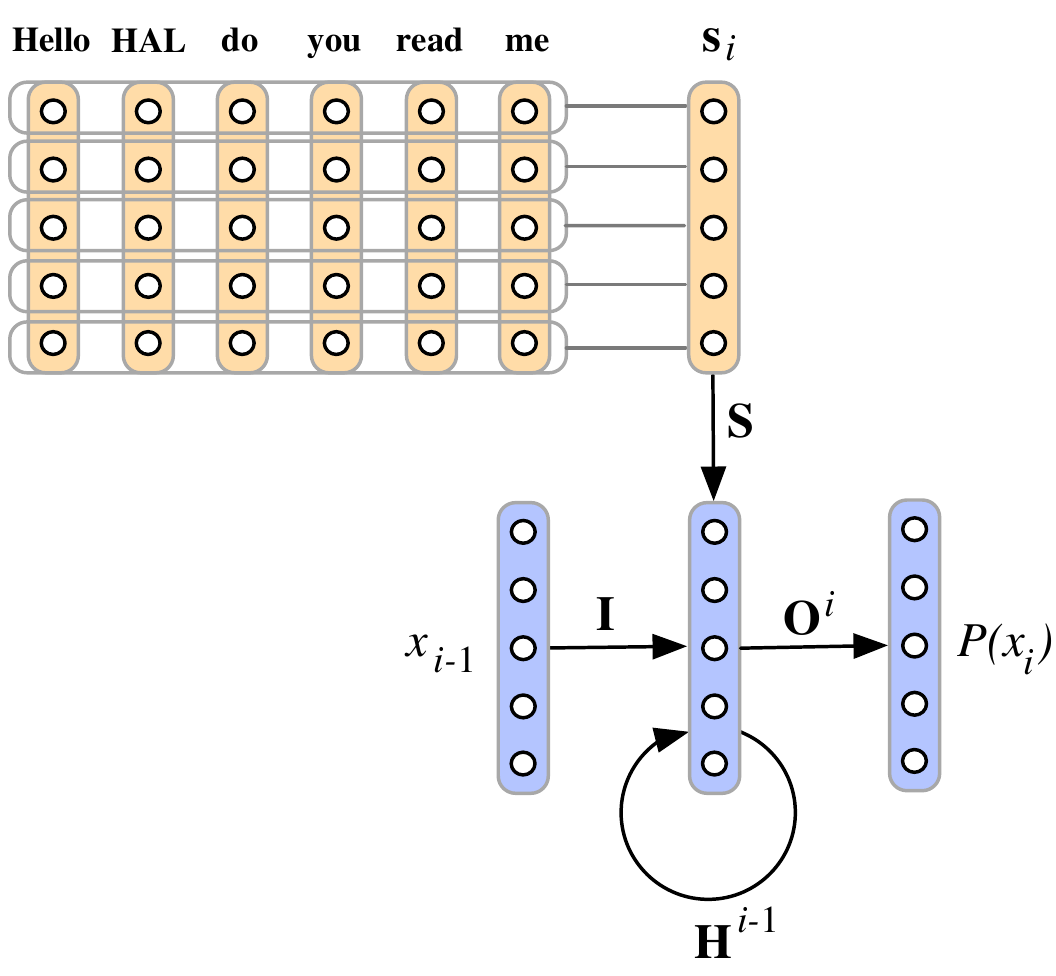}

\caption{Recurrent convolutional neural network (RCNN) discourse model based on a RNN architecture. At each step the RCNN takes as input the current sentence vector $\vec{s}_i$ generated through the HCNN sentence model and the previous label $x_{i-1}$ to predict a probability distribution over the current label $P(x_i)$. The recurrent weights $\vec{H}^{i-1}$ are conditioned on the previous agent $a_{i-1}$ and the output weights are conditioned on the current agent $a_i$. Note also the sentence matrix $\vec{M}_s$ of the sentence model and the hierarchy of convolutions applied to each feature that is a row in $\vec{M}_s$ to produce the corresponding feature in $\vec{s}_i$. }
\end{figure}
\subsection{Sentential compositionality}

There are three main aspects of sentential compositionality that the model aims at capturing. To relate these, it is useful to note the following basic property of the model: a sentence $s$ is paired to the matrix $\vec{M}^s$ whose columns are given sequentially by the vectors of the words in $s$. A row in $\vec{M}^s$ corresponds to the values of the corresponding feature across all the word vectors. The first layer of the network in Fig. 1 represents one such row of $\vec{M}^s$, whereas the whole matrix $\vec{M}^s$ is depicted in Fig. 2. The three considerations are as follows.

First, at the initial stage of the composition, the value of a feature in the sentence vector is a function of the values of the same feature in the word vectors. That is, the $m$-th value in the sentence vector of $s$ is a function of the $m$-th row of $\vec{M}^s$. This aspect is preserved in the additive and multiplicative models where the composition operations are, respectively, addition $+$ and component-wise multiplication $\odot$. 
The current model preserves the aspect up to the computation of the sentence vector $\vec{s}$ by adopting one-dimensional, feature-wise convolution operations.  Subsequently, the discourse model that uses the sentence vector $\vec{s}$ includes transformations across the features of $\vec{s}$ (the transformation $\vec{S}$ in Fig. 2). 

The second consideration concerns the {hierarchical} aspect of the composition operation. We take the compositionality of meaning to initially yield local effects across neighbouring words and then yield increasingly more global effects across all the words in the sentence. Composition operations like those in the structured models that are guided by the syntactic parse tree of the sentence capture this trait. 
The sentence model preserves this aspect not by way of syntactic structure, but by adopting convolution kernels of {gradually increasing sizes} that span an increasing number of words and ultimately the entire sentence.

The third aspect concerns the dependence of the composition operation. The operation is taken to depend on the different features, but not on the different words. Word specific parameters are introduced only by way of the learnt word vectors, but no word specific operations are learnt. We achieve this by using a single convolution kernel across a feature, and by utilizing different convolution kernels for different features.
Given these three aspects of sentential compositionality, we now proceed to describe the sentence model in detail.

\subsection{Hierarchical Convolutional Neural Network}
The sentence model is taken to be a CNN where the convolution operation is applied one dimensionally across a single feature and in a hierarchical manner. 
To describe it in more detail, we first recall the convolution operation that is central to the model. Then we describe how we compute the sequence of kernel sizes and how we determine the hierarchy of layers in the network.

\subsubsection{Kernel and One-dimensional Convolution}
Given a sentence $s$ and its paired matrix $\vec{M}^s$, let $\vec{m}$ be a feature that is a row in $\vec{M}^s$. Before defining kernels and the convolution operation, let us consider the underlying operation of \emph{local weighted addition}. Let $w_1, ..., w_k$ be a sequence of $k$ weights; given the feature $\vec{m}$, local weighted addition over the \emph{first} $k$ values of $\vec{m}$ gives:
\begin{equation}y = w_1\vec{m}_1 + ... + w_k\vec{m}_k\end{equation}
Then, a kernel simply defines the value of $k$ by specifying the sequence of weights $w_1,...,w_k$ and the one-dimensional convolution applies local weighted addition with the $k$ weights to each subsequence of values of $\vec{m}$.

More precisely, let a one-dimensional \emph{kernel} $\vec{k}$ be a vector of weights and assume $|\vec{k}| \leq |\vec{m}|$, where $|\cdot|$ is the number of elements in a vector. Then we define the discrete, valid, \emph{one-dimensional convolution} $(\vec{k} * \vec{m})$ of kernel $\vec{k}$ and feature $\vec{m}$ by: 
\begin{equation}
(\vec{k} * \vec{m})_i := \sum_{j=1}^k \vec{k}_j \cdot \vec{m}_{k+i-j}
\end{equation}
where $k = |\vec{k}|$ and 
$|\vec{k} * \vec{m}| = |\vec{m}|-k+1$. Each value in $\vec{k} * \vec{m}$ is a sum of $k$ values of $\vec{m}$ weighted by values in $\vec{k}$ (Fig. 3). To define the hierarchical architecture of the model, we need to define a sequence of kernel sizes and associated weights. To this we turn next.
\begin{figure}
\centering
\includegraphics[width=0.35\textwidth]
{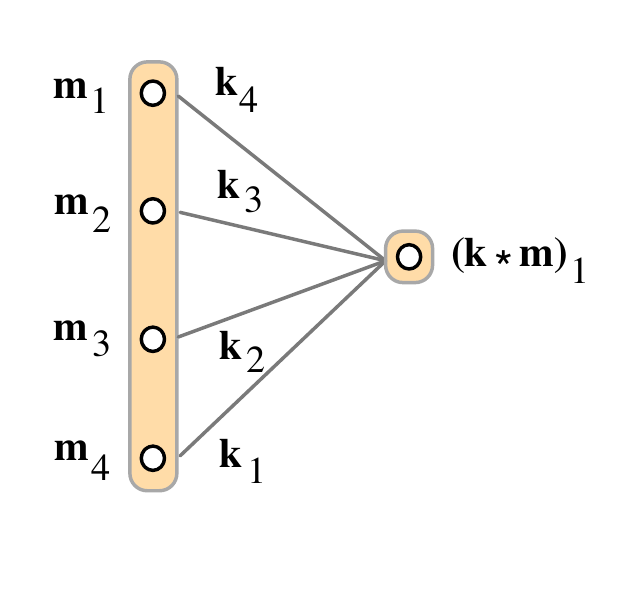}
\caption{Convolution of a vector $\vec{m}$ with a kernel $\vec{k}$ of size 4.}
\end{figure}
\subsubsection{Sequence of Kernel Sizes}

Let $l$ be the number of words in the sentence $s$. The sequence of kernel sizes $\langle k^l_i\rangle_{i\leq t}$ depends only on the length of $s$ and itself  has length $t = \lceil\sqrt{2l}\rceil -1$. It is given recursively by:
\begin{equation}
k^l_1 = 2, \quad
k^l_{i+1} = k^l_{i}+1, \quad 
k^l_{t} = l - \sum_{j=1}^{t-1} (k^l_j-1)
\end{equation}
That is, kernel sizes increase by one until the resulting convolved vector is smaller or equal to the last kernel size; see for example the kernel sizes in Fig. 1. Note that, for a sentence of length $l$, the number of layers  in the HCNN including the input layer will be $t+1$ as convolution with the corresponding kernel is applied at every layer of the model. Let us now proceed to define the hierarchy of layers in the HCNN.
\begin{figure*}
\includegraphics[width=1\textwidth]{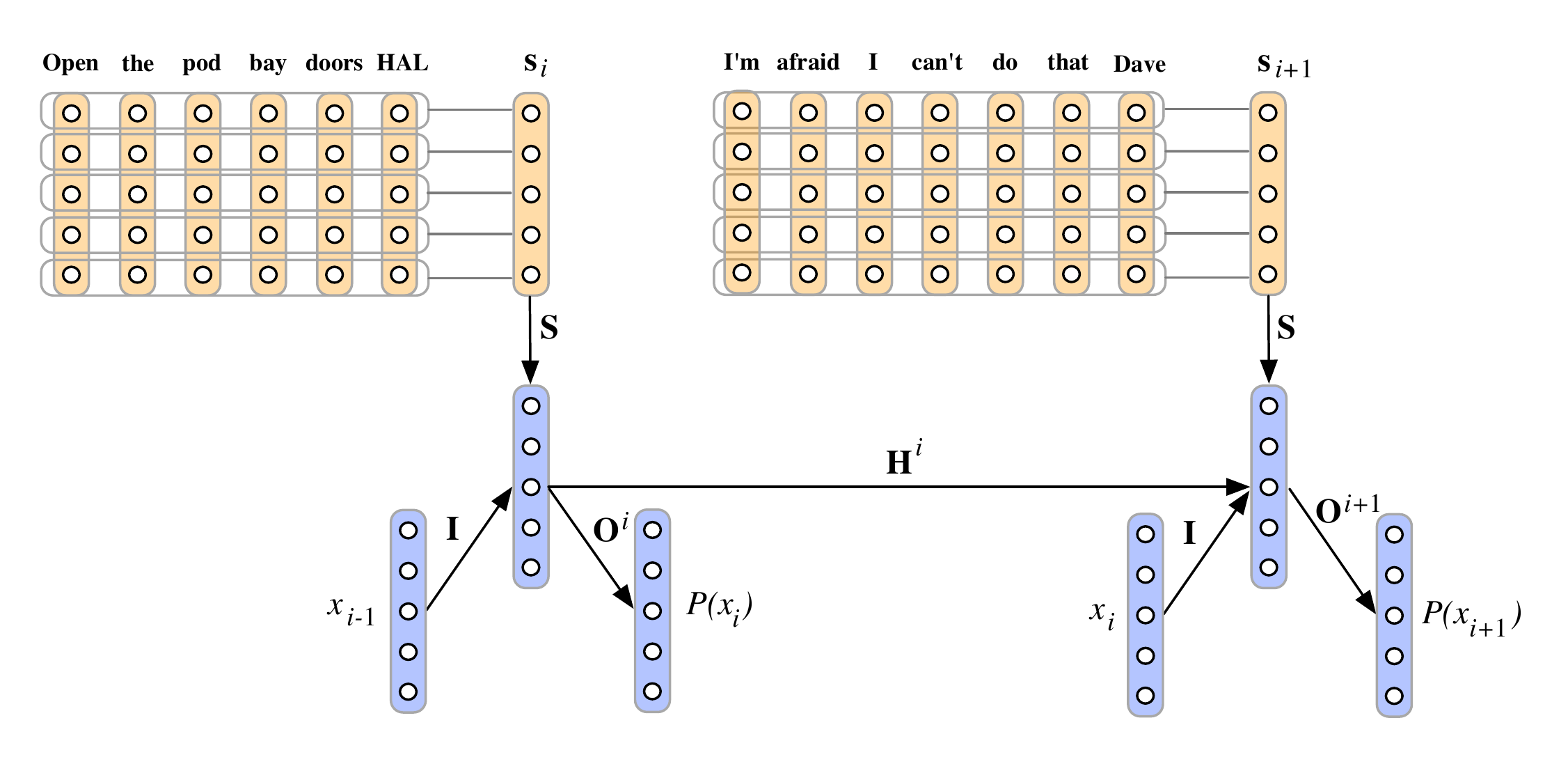}

\caption{Unravelling of a RCNN discourse model to depth $d=2$. The recurrent $\vec{H}^i$ and output $\vec{O}^i$ weights are conditioned on the respective agents $a_i$.}
\end{figure*}

\subsubsection{Composition Operation in a HCNN}

Given a sentence $s$, its length $l$ and a sequence of kernel sizes $\langle k^l_i\rangle_{i\leq t}$, we may now give the recursive definition that yields the hierarchy of one-dimensional convolution operations applied to each feature $f$ that is a row in $\vec{M}^s$. Specifically, for each feature $f$, let $\vec{K}^f_i$ be a sequence of $t$ kernels, where the size of the kernel $|\vec{K}^f_i| = k_i^l$. Then we have the hierarchy of matrices and corresponding features as follows:
\begin{align}
\vec{M}^1_{f,:} &= \vec{M}^s_{f,:} \\
\vec{M}^{i+1}_{f,:} &= \sigma(\ \vec{K}_i^f*\vec{M}^{i}_{f,:} + b_f^i \ )
\end{align}
for some non-linear sigmoid function $\sigma$ and bias $b_f^i$, where $i$ ranges over $1,...,t$.
In sum, one-dimensional convolution is applied feature-wise to each feature of a matrix at a certain layer, where the kernel weights depend both on the layer and the feature at hand (Fig. 1). A hierarchy of matrices is thus generated with the top matrix being a single vector for the sentence.

\subsubsection{Multiple merged HCNNs}

Optionally one may consider multiple parallel HCNNs that are merged according to different strategies either at the top sentence vector layer or at intermediate layers. The weights in the word vectors may be tied across different HCNNs. Although potentially useful, multiple merged HCNNs are not used in the experiment below.

This concludes the description of the sentence model.
Let us now proceed to the discourse model.

\newpage

\section{ Discourse Model}

The discourse model adapts a RNN architecture in order to capture central properties of discourse. We here first describe such properties and then define the model itself.

\subsection{Discourse Compositionality}

The meaning of discourse - and of words and utterances within it - is often a result of a rich ensemble of context, of speakers' intentions and actions and of other relevant surrounding circumstances \cite{sep-pragmatics,Potts2011PRAGMATICS}. Far from capturing all aspects of discourse meaning, we aim at capturing in the model at least two of the most prominent ones: the sequentiality of the utterances and the interactions between the speakers.

Concerning sequentiality, just the way the meaning of a sentence generally changes if words in it are permuted, so does the meaning of a paragraph or dialogue change if one permutes the sentences or utterances within. The change of meaning is more marked the larger the shift in the order of the sentences. Especially in tasks where one is concerned with a specific sentence within the context of the previous discourse, capturing the order of the sentences preceding the one at hand may be particularly crucial.

Concerning the speakers' interactions, the meaning of a speaker's utterance within a discourse is differentially affected by the speaker's previous utterances as opposed to \emph{other} speakers' previous utterances. Where applicable we aim at making the computed meaning vectors reflect the current speaker and the sequence of interactions with the previous speakers. With these two aims in mind, let us now proceed to define the model.

\begin{table*}
\centering
\begin{tabular}[t]{| l l c c |} \hline
\text{Dialogue Act Label} & \text{Example} & \text{Train (\%)} & \text{Test (\%)} \\ \hline
$\mathsf{Statement}$ & \emph{And, uh, it's a legal firm office.} & 36.9 & 31.5 \\
$\mathsf{Backchannel/Acknowledge}$ & \emph{Yeah, anything could happen.} & 18.8 & 18.2 \\
$\mathsf{Opinion}$ &\emph{I think that would be great.} & 12.7& 17.1\\
$\mathsf{Abandoned/Uninterpretable}$   & \emph{So, -} & 7.6&8.6\\
$\mathsf{Agreement/Accept}$& \emph{Yes, exactly.}  & 5.5 & 5.0\\
$\mathsf{Appreciation}$&  \emph{Wow.} & 2.3& 2.2 \\
$\mathsf{Yes\! -\! No\! -\! Question}$ &  \emph{Is that what you do?} & 2.3& 2.0\\
$\mathsf{Non\!-\!Verbal}$& \emph{[Laughter], [Throat-clearing]}  & 1.7 & 1.9 \\ \hline
\emph{Other labels (34)} & & 12.2 & 13.5\\ \hline\hline
\emph{Total number of utterances} & & 196258 & 4186 \\ 
\emph{Total number of dialogues} & & 1115 & 19 \\ \hline
\end{tabular}

\caption{Most frequent dialogue act labels with examples and frequencies in train and test data.}
\end{table*}

\subsection{Recurrent Convolutional Neural Network}
The discourse model coupled to the sentence model is based on a RNN architecture with inputs from a HCNN and with the recurrent and output weights conditioned on the respective speakers.

We take as given a sequence of  sentences or utterances $s_1,...,s_T$, each in turn being a sequence of words $s_i = y_1^i...y_l^i$, a sequence of labels $x_1,...,x_T$ and a sequence of speakers or agents $a_1, ..., a_T$, in such way that the $i$-th utterance is performed by the $i$-th agent and has label $x_i$. We denote by $\vec{s}_i$ the sentence vector computed by way of the sentence model for the sentence $s_i$. The RCNN computes probability distributions $p_i$ for the label at step $i$ by iterating the following equations:
\begin{align}
\vec{h}_i &= \sigma(\ \vec{I} x_{i-1} + \vec{H}^{{i-1}}\vec{h}_{i-1} + \vec{S}  \vec{s}_i  + b_h ) \\
p_i &= softmax( \vec{O}^{i}\vec{h}_i + b_o)
\end{align}
where $\vec{I}, \vec{H}^i, \vec{O}^i$ are corresponding weight matrices for each agent $a_i$ and $softmax(y)_k = \displaystyle \frac{e^{y_k}}{ \sum_j e^{y_j}}$ returns a probability distribution. Thus $p_i$ is taken to model the following predictive distribution:
\begin{equation}
p_i = P(x_i|x_{<i},s_{\leq i}, a_{\leq i})
\end{equation}
An RCNN and the unravelling to depth $d=2$ are depicted respectively in Fig. 2 and Fig. 4. With regards to vector representations of discourse, we take the hidden layer $\vec{h}_i$ as the vector representing the discourse up to step $i$. This concludes the description of the discourse model. Let us now consider the experiment.

\section{Predicting Dialogue Acts}

We experiment with the prediction of dialogue acts within a conversation. A dialogue act specifies the pragmatic role of an utterance and helps identifying the speaker's intentions \cite{Austin,sep-pragmatics}. The automated recognition of dialogue acts is crucial for dialogue state tracking within spoken dialogue systems \cite{Williams:2012:BTC:2390444.2390461}. We first describe the Switchboard Dialogue Act (SwDA) corpus \cite{0a0045064fb44dcc957522a58fd75771} that serves as the dataset in the experiment. We report on the training procedure and the results and we make some qualitative observations regarding the discourse representations produced by the model. 
\begin{table*}
\centering
\small
\begin{tabular}[t]{| l || l || l || l |} \hline
Center & A: \emph{Do you repair your own car?} & A: \emph{-- I guess we can start.} & A: \emph{Did you use to live around here?} \\
Dialogue& B: \emph{I try to, whenever I can.} & B: \emph{Okay.} & B: \emph{Uh, Redwood City.}  \\ \hline
First NN & A: \emph{Do you do it every day?} & A: \emph{I think for serial murder --}  & A: \emph{Can you stand up in it?} \\
& B: \emph{I try to every day.}  & B: \emph{Uh-huh.} & B: \emph{Uh, in parts.} \\ \hline
Second NN  & A: \emph{Well, do you have any children?}& A: \emph{The USSR -- wouldn't do it } & A: \emph{[Laughter]  Do you have any kids} \\
& & & \emph{ that you take fishing?}   \\
& B: \emph{I've got one.} & B: \emph{Uh-huh.}  & B: \emph{Uh, got a stepdaughter.}  \\ \hline
Third NN & A: \emph{Do you manage the money?} & A: \emph{It seems to me there needs}  & A: \emph{Is our five minutes up?} \\ 
&  & \emph{to be some ground, you know,} & \\
& & \emph{some rules --} &  \\
& B: \emph{Well, I, we talk about it.} & B: \emph{Uh-huh.}  & B: \emph{Uh, pretty close to it.} \\ \hline
Fourth NN & A: \emph{ Um, do you watch it every} & A: \emph{It sounds to me like, uh,} & A: \emph{Do you usually go out, uh,}  \\
& \emph{Sunday?} & \emph{you are doing well.} & \emph{with the children or without them?} \\
 & B: \emph{[Breathing]  Uh, when I can.} & B: \emph{My husband's retired.} & B: \emph{Well, a variety.}\\ \hline
\end{tabular}

\caption{Short dialogues and nearest neighbours (NN).}
\end{table*}
\subsection{SwDA Corpus}

The SwDA corpus contains audio recordings and transcripts of telephone conversations between multiple speakers that do not know each other and are given a topic for discussion. For a given utterance we use the transcript of the utterance, the dialogue act label and the speaker's label; no other annotations are used in the model. Overall there are 42 distinct dialogue act labels such as $\mathsf{Statement}$ and $\mathsf{Opinion}$ (Tab.\!\! 1). We adopt the same data split of 1115 train dialogues and 19 test dialogues as used in \cite{DBLP:journals/coling/StolckeRCSBJTMM00}.

\subsection{Objective Function and Training}
We minimise the cross-entropy error of the predicted and the true distributions and include an $l_2$ regularisation parameter. The RCNN is truncated to a  depth $d=2$ so that the prediction of a dialogue act depends on the previous two utterances, speakers and dialogue acts; adopting depths $>2$ has not yielded improvements in the experiment. The derivatives are efficiently computed by back-propagation \cite{RumHinWill}. The word vectors are initialised to random vectors of length 25 and no pretraining procedure is performed. We minimise the objective using L-BFGS in mini-batch mode; the minimisation converges smoothly.

\subsection{Prediction Method and Results}

The prediction of a dialogue act is performed in a greedy fashion. Given the two previously predicted acts $\hat{x}_{i-1},\hat{x}_{i-2}$, one chooses the act $\hat{x_i}$ that has the maximal probability in the predicted distribution $P(x_i)$. The LM-HMM model of \protect\cite{DBLP:journals/coling/StolckeRCSBJTMM00} learns a language model for each dialogue act and a Hidden Markov Model for the sequence of dialogue acts and it requires all the utterances in a dialogue in order to predict the dialogue act of any one of the utterances. The RCNN makes the weaker assumption that only the utterances up to utterance $i$ are available to predict the dialogue act $\hat{x_i}$. The accuracy results of the models are compared in {Tab.\,3}.
\begin{table}
\centering
\begin{tabular}{| r | c |} \hline
 & Accuracy (\%)  \\ \hline
   RCNN & \textbf{73.9}   \\
   LM-HMM trigram & 71.0 \\
   LM-HMM bigram & 70.6\\
	LM-HMM unigram & 68.2\\
  Majority baseline & 31.5 \\
  Random baseline & 2.4\\\hline
\end{tabular}
\caption{SwDA dialogue act tagging accuracies. The LM-HMM results are from \protect\cite{DBLP:journals/coling/StolckeRCSBJTMM00}. Inter-annotator agreement and theoretical maximum is 84\%.}
\end{table}

\subsection{Discourse Vector Representations}

We inspect the discourse vector representations that the model generates. After a dialogue is processed, the hidden layer $\vec{h}$ of the RCNN is taken to be the vector representation for the dialogue (Sect. 3.2). Table 2 includes three randomly chosen dialogues composed of two utterances each; for each dialogue the table reports the four nearest neighbours. As the word vectors and weights are initialised randomly without pretraining, the word vectors and the weights are induced during training only through the dialogue act labels attached to the utterances. The distance between two word, sentence or discourse vectors reflects a notion of pragmatic similarity: two words, sentences or discourses are similar if they contribute in a similar way to the pragmatic role of the utterance signalled by the associated dialogue act. This is suggested by the examples in Tab. 2, where a centre dialogue and a nearest neighbour may have some semantically different components (e.g. ``repair your own car" and  ``manage the money"), but be pragmatically similar and the latter similarity is captured by the representations. In the examples, the meaning of the relevant words in the utterances, the speakers' interactions and the sequence of pragmatic roles are well preserved across the nearest neighbours.

\section{Conclusion}

Motivated by the compositionality of meaning both in sentences and in general discourse, we have introduced a sentence model based on a novel convolutional architecture and a discourse model based on a novel use of recurrent networks. We have shown that the discourse model together with the sentence model achieves state of the art results in a dialogue act classification experiment without feature engineering or pretraining and with simple greedy decoding of the output sequence. We have also seen that the discourse model produces compelling discourse vector representations that are sensitive to the structure of the discourse and promise to capture subtle aspects of discourse comprehension, especially when coupled to further semantic data and unsupervised pretraining.

\section*{Acknowledgments}

We thank Ed Grefenstette and Karl Moritz Hermann for great conversations on the matter.
The authors gratefully acknowledge the support of the Clarendon Fund and the EPSRC.

\bibliographystyle{acl}
\bibliography{Master}

\end{document}